\documentclass{article}

\usepackage[preprint]{neurips_2026}

\usepackage[utf8]{inputenc} 
\usepackage[T1]{fontenc}    
\usepackage{hyperref}       
\usepackage{url}            
\usepackage{booktabs}       
\usepackage{amsfonts}       
\usepackage{nicefrac}       
\usepackage{microtype}      
\usepackage{xcolor}         
\usepackage{graphicx}
\usepackage{amsmath}
\usepackage{makecell}
\usepackage{multirow}
\usepackage{subcaption}

\title{Explicit Kinematic Guidance from Analytic Concepts for Vision-Language-Action Models}

\author{
  Mingyang Sun$^{1,2,4}$
  \quad
  Jiude Wei$^{5}$
  \quad
  Xiujian Liang$^{2,3}$
  \quad
  Qichen He$^{2,5}$
  \\
  \textbf{Donglin Wang}$^{4}$
  \quad
  \textbf{Cewu Lu}$^{2,5}$
  \quad
  \textbf{Jianhua Sun}$^{5}$\thanks{Corresponding author.}
  \\[0.5em]
  $^{1}$Zhejiang University, Hangzhou, China \quad
  $^{2}$Shanghai Innovation Institute, Shanghai, China \\
  $^{3}$Fudan University, Shanghai, China \quad
  $^{4}$Westlake University, Hangzhou, China \\
  $^{5}$Shanghai Jiao Tong University, Shanghai, China
  \\[0.2em]
  \small{\href{mailto:sunmingyang@zju.edu.cn}{\texttt{sunmingyang@zju.edu.cn}},\quad \href{mailto:gothic@sjtu.edu.cn}{\texttt{gothic@sjtu.edu.cn}}}
}

\begin{document}

\maketitle

\begin{abstract}
Current Vision-Language-Action (VLA) models rely mainly on 2D inputs, neglecting the rich object structural information and commonsense knowledge inherent in the 3D physical world. This deficiency restricts their spatial awareness and adaptability for complex, high-precision manipulation. To bridge this crucial gap, we construct a Concept Expert module for VLA to build executable Analytic Concepts that represent objects as explicit, programmatic blueprints. Our mechanism operates in two synergistic phases: First, prior to VLA inference, the Concept Expert leverages 3D information from Vision Foundation Models (VFMs) to estimate the initial kinematic and structural parameters. Second, throughout the manipulation process, the VLA model utilizes its inherent capability to dynamically track the dynamic concept parameters, continuously aligning them with observational changes to ensure persistent accuracy. Once established, the Analytic Concepts provide explicit, high-quality guidance for VLA fine-tuning through (1) dense, programmatic manipulation rewards and (2) precise spatial guidance.
  This formulation allows VLA models to learn physically grounded interaction behaviors while maintaining end-to-end learning flexibility.
 Our experimental results show consistent improvements in success rate and learning efficiency across supervised and reinforcement learning settings, demonstrating the effectiveness of structured, concept-based guidance for VLA post-training.
 Project page: \href{https://sunmmyy.github.io/sage/}{\texttt{sunmmyy.github.io/sage/}}.
  
\end{abstract}

\section{Introduction}
\label{sec:intro}




Robots that can seamlessly follow human instructions and manipulate unfamiliar objects in unstructured 3D environments remain an open challenge. Recent progress in Vision-Language-Action (VLA) models has brought this ambition closer to reality by enabling end-to-end mapping from sensory inputs and natural language instructions to robot actions~\citep{DBLP:conf/corl/ZitkovichYXXXXW23,sapkota2025visionlanguageactionmodelsconceptsprogress,embodimentcollaboration2025openxembodimentroboticlearning}. However, their reliance on vast amounts of data and their ability to generalize remain significantly limited. Even slight changes in the environment, such as variations in object appearance or illumination, can dramatically degrade performance. Consequently, fine-tuning these VLA models for downstream tasks remains data-intensive, often requiring substantial amounts of task- and environment-specific data to adapt to the target visual and action domains. We argue that this limitation stems from a fundamental representational gap: VLA models trained primarily on 2D inputs often overlook the structured abstractions and commonsense knowledge in 3D scenes that capture the fundamental patterns underlying robotic manipulation~\citep{li2025spatialforcingimplicitspatial}. Without such geometric and physical priors, the policy must rediscover the same commonsense constraints—e.g., doors rotate about hinges, drawers translate along rails—every time it encounters a new scene layout, squandering data and computation.

Cognitive science offers a clue. The \emph{Recognition-by-Components} theory~\citep{Biederman1987RecognitionbycomponentsAT} posits that humans recognise and manipulate objects by decomposing them into elementary volumetric primitives and their spatial relations. Inspired by this idea, we adopt the Analytic Concept annotation system~\citep{DBLP:conf/nips/0003LXWWZL24,sun2025discovering}, which offer a promising direction by representing structural knowledge through explicit, programmatic blueprints. These blueprints effectively encapsulate the topological structures and affordance knowledge crucial for physical interaction. By leveraging this structural representation, our approach moves beyond passive 2D perception to integrate active object-centric spatial understanding. This conceptual grounding provides the necessary scaffolding to guide VLA models, enhancing their ability to reason about physics and structure, thereby improving generalization and fine-tuning efficiency.

To effectively leverage the structural richness of Analytic Concepts and address the limitations of 2D-centric VLA models, we propose SAGE (\underline{S}patial \underline{A}nalytic-concept \underline{G}uided \underline{E}nhancement), a novel fine-tuning framework that injects explicit spatial and physical guidance into the VLA training loop. SAGE wraps any existing VLA model with a \emph{Concept Expert} to instantiate the relevant Analytic Concept before each episode begins. Two synergistic components designed in the concept expert to facilitate robust, object-centric learning:
\begin{itemize}
    \item Prior to VLA inference, the expert is responsible for processing 3D information derived from Vision Foundation Models (VFMs)~\citep{VGGT,SAM2} to estimate the initial kinematic and structural parameters of the Analytic Concepts.
    \item To maintain the fidelity of these structural priors in dynamic environments, we allows the VLA model itself to assist in the dynamic tracking of these concept parameters amidst continuous observation changes.
\end{itemize}
This two-pronged approach ensures both high-quality initialization and consistent tracking of structural knowledge. Executing the resulting blueprint on-the-fly yields two complementary training signals: (1) dense, programmatic manipulation rewards for Reinforcement Learning~(RL) finetuning, and (2) precise, low-variance guidance vectors that can be consumed by the action expert for finetuning.  At deployment time, the learned policy runs directly on raw camera observations; only a lightweight forward pass through the Concept Expert is required. 
In summary, this work makes the following four main contributions:
\begin{itemize}
    \item We present a new paradigm for VLA fine-tuning that augments policy learning with executable object-centric structural priors from Analytic Concepts.
    \item We introduce a Concept Expert that instantiates and tracks concept parameters from visual observations, enabling dynamic alignment between object structure and policy execution.
    \item We develop a unified guidance mechanism that translates Analytic Concepts into action-space constraint regularization and dense geometry-aware rewards for policy optimization.
    \item We demonstrate consistent improvements in performance and sample efficiency across a range of manipulation tasks, particularly those involving articulated objects.
\end{itemize}


\section{Preliminaries}

\subsection{Problem Setup}
We formulate each robotic task as a \textbf{Markov Decision Process (MDP)}, defined as $\mathcal{M} = (\mathcal{S}, \mathcal{A}, \mathcal{P}, r, \rho_0, \gamma)$, where $\mathcal{S}$ and $\mathcal{A}$ denote the state and action spaces, respectively. The environment transition dynamics are characterized by conditional probabilities $\mathcal{P}(\boldsymbol{s}'|\boldsymbol{s},\boldsymbol{a})$ governed by the system dynamics, and $\rho_0(\boldsymbol{s})$ specifies the initial state distribution. The reward function is represented by $r(\boldsymbol{s},\boldsymbol{a})$, with $\gamma \in (0,1)$ as the discount factor. The environment provides state transitions $\mathcal{P}$ and reward signals as follows:

\begin{align}
r_t = \alpha \cdot I_{\text{success}} + (1 - \alpha) \cdot \sum_i w_i \cdot \phi_i(\boldsymbol{s}_t, \boldsymbol{a}_t),
\alpha \in [0,1], \quad
I_{\text{success}} = \begin{cases}
1, & \text{if task success} \\
0, & \text{otherwise}
\end{cases}.
\end{align}
Here, $I_{\text{success}}$ is an indicator function signaling task success, and $\{\phi_i\}$ are auxiliary reward features with corresponding weights $\{w_i\}$.

Our work focuses on fine-tuning a pre-trained vision-language-action (VLA) model for downstream robotic tasks. We assume access to a pre-trained VLA model $\pi_{\phi_{\text{pre}}}$, which encodes high-level representations from multimodal inputs including visual observations $\boldsymbol{o}^{\text{vis}}_{t}$ (e.g., RGB images), proprioceptive states $\boldsymbol{o}^{\text{prop}}_t$ (e.g., joint angles, end-effector pose), and language instructions. Formally, the state is defined as:
\begin{equation}
    \boldsymbol{s}_t = (\boldsymbol{o}^{\text{vis}}_{t}, \boldsymbol{o}^{\text{prop}}_t, l_{\text{task}}),
\end{equation}

In Supervised Fine-Tuning (SFT), the goal is to adapt the pre-trained parameters $\phi_{\text{pre}}$ to the target task using a limited set of labeled demonstrations. Formally, given a demonstration trajectory $\tau = (\boldsymbol{s}_0, \boldsymbol{a}_0, \ldots, \boldsymbol{s}_H, \boldsymbol{a}_H)$, the fine-tuning objective is to solve: $\min_{\phi} \mathcal{L}(\tau, \phi),$ where the loss function $\mathcal{L}$ can be defined as the negative log-likelihood (NLL) or mean squared error (MSE) between predicted and demonstrated actions, measuring the discrepancy and thereby enabling effective policy adaptation. The overarching goal of Reinforcement Learning (RL) is to learn a policy $\pi(\boldsymbol{a}|\boldsymbol{s})$ that maximizes the expected return, defined as the discounted sum of future rewards:
$G_t = \sum_{i=0}^\infty \gamma^i r_{t+i}$.



\begin{figure*}
   \centering
  \includegraphics[trim = 23mm 60mm 13mm 35mm, clip, width=0.99
  \linewidth]{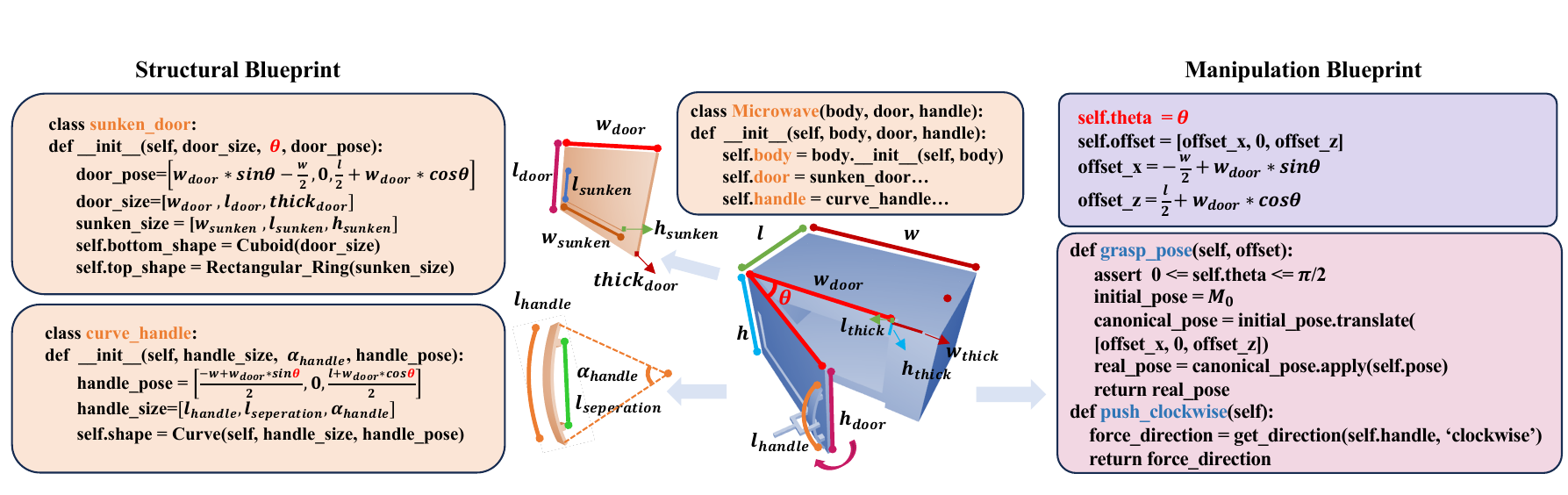}
  \caption{Example implementation of Analytic Concepts. Structural Blueprint: higher-level objects are procedurally composed by wiring multiple geometric assets together, forming a parametric graph that captures their spatial layout and structural relationships. Manipulation Blueprint: parameterised routines compute grasp poses and force directions that exploit the affordances encoded in the underlying structure.}
  \label{fig:AC}
\end{figure*}

\subsection{Analytic Concepts}


The foundation of our work is the notion of \textbf{Analytic Concepts} (ACs), a previously proposed annotation paradigm for abstracting structural knowledge of the physical world~\citep{DBLP:conf/nips/0003LXWWZL24}. ACs represent regular geometric patterns and their associated interaction knowledge as explicit, reusable concepts. With the assistance of VLMs, this paradigm can support automatic concept annotation by identifying object components, geometric relations, and affordance-relevant structures, thereby enabling structural knowledge to be propagated across object instances. Prior work has validated this paradigm on approximately 4,400 objects; additional details are provided in Appendix~\ref{sec:ACS}.

At its foundation, this paradigm provides a library of geometric \textbf{concept assets}~(e.g., cube and sphere). These assets serve as the atomic building blocks, each providing (1) free parameters for variations, (2) a canonical structural definition, and (3) affordance annotations. These assets are assembled into: \textbf{Structural Blueprints:} Mathematical procedures defining the common spatial structure (layout, relationships) shared by instances (Fig.~\ref{fig:AC}left); \textbf{Manipulation Blueprints:} Procedures defining how the object can be interacted with (Fig.~\ref{fig:AC}right).

Specifically, the concept parameters $P$ are categorized into:
\begin{itemize}
    \item \textbf{Structural Parameters ($P_{s}$):} Time-invariant properties that define the object's geometry and kinematic structure (e.g., the axis of rotation for a hinge, or the direction and extent of a sliding track).
    
    \item \textbf{Kinematic Parameters ($P_{k}$):} Time-varying properties that describe the current state of the object's movable components (e.g., the displacement of a drawer along its track, or the rotation angle of a knob).
\end{itemize}
This system provides a rich 3D dataset of concept instances, which crucially enables the {pre-trained parameter estimation networks}. This pre-trained model (which we later utilize as our Concept Expert) can accurately resolve a target object in the physical world into its corresponding parameters $\{P_{s}, P_{k}\}$.
This programmatic structure allows the VLA model to reason about object affordances based on verifiable physical blueprints.
\section{Methodology}

\begin{figure*}[tb]
   \centering
  \includegraphics[trim = 0mm 0mm 0mm 0mm, clip, width=0.99
  \linewidth]{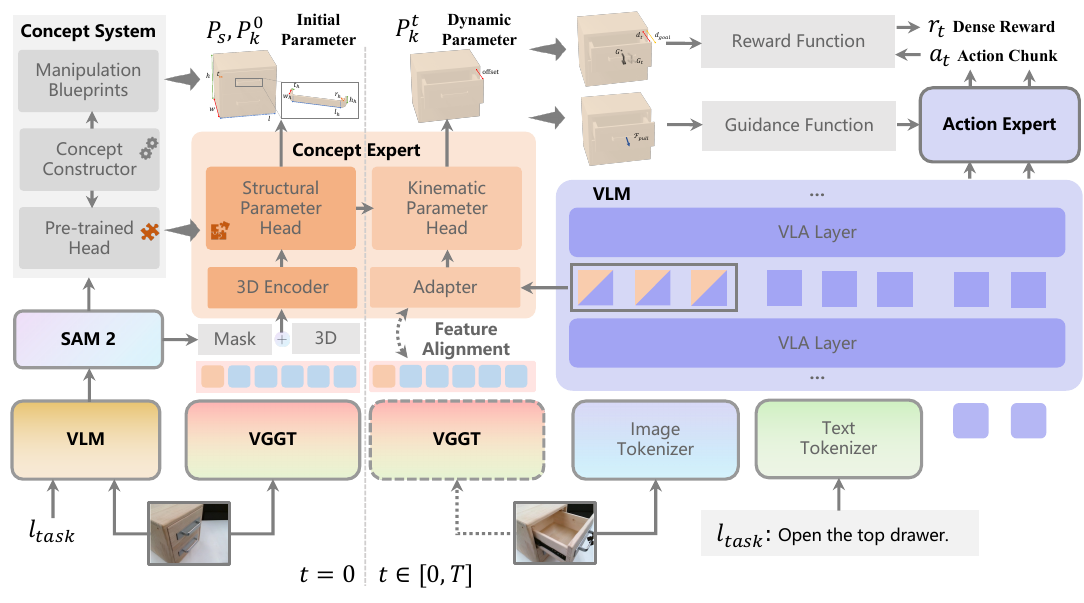}
  \caption{SAGE Architecture Overview. At $t=0$ (Initialization Phase, left): The Concept Expert utilizes 3D features from VGGT~\citep{VGGT} and SAM~\citep{SAM2} to estimate the initial structural parameters ($P_s$) and kinematic parameters ($P_k^0$) via the Structural Parameter Head. For $t>0$ (Dynamic Tracking and Rollout Phase, right): An adapter extracts intermediate features from the VLA model's VLA Layer. This enables the Concept Expert's Dynamic Parameter Head to track the dynamic parameters ($P_k^t$) in real-time. The constructed analytic concept can use Reward Function and Guidance Function to generate dense feedback signals (like $r_t$) for fine-tuning the Action Expert. Dashed lines indicate training-only.}
  \label{fig:overview}
\end{figure*}

In this section, we introduce SAGE-VLA, our novel framework designed to inject explicit structural knowledge into VLA fine-tuning. We first provide an overview of the SAGE architecture, followed by a detailed description of its core components: the Concept Expert module (including analytic concept initialization and the dynamic parameter tracking),  and the mechanism for generating explicit guidance signals .

\subsection{Overview}

The primary objective of the SAGE framework is to seamlessly integrate the robust structural and physical understanding inherent in Analytic Concepts into the VLA training pipeline. Our core contribution is to transform the implicit spatial reasoning within the VLA model into explicit, high-quality, knowledge-driven guidance. For this, the SAGE architecture introduces the Concept Expert module as an auxiliary, structured knowledge stream that tightly integrates with the main VLA training pipeline, as illustrated in Fig.~\ref{fig:overview}. The Concept Expert module serves two critical roles: 
\begin{enumerate}
    \item \textbf{Initialization ($t=0$):} It estimates the initial structural parameters ($P_s$) and kinematic state ($P_k^0$) using dedicated 3D processing, providing an accurate spatial anchor;
    \item  \textbf{Dynamic Tracking ($t \in [0, T]$):} It actively tracks the object's kinematic state ($P_k^t$) by exploiting a Feature Alignment mechanism, which bridges the VLA's internal features with the Concept Expert, allowing the VLA's perception to assist in the tracking process.
\end{enumerate}

The dynamically tracked parameters ($P_k^t$) are then utilized by Reward and Guidance functions to provide dense, structured feedback to the VLA's output layers (Action Expert), thereby refining the VLA's policy ($\pi$) to produce geometrically grounded actions ($\boldsymbol{a}_t$).

\subsection{Analytic Concept Initialization}

The first phase of our method is to establish a high-fidelity, structural understanding of the scene at $t=0$. This structural initialization is handled by the Concept Expert and provides the foundational parameters $P_s$ (Structural) and $P_k^0$ (Initial Kinematic).

As illustrated in Fig.~\ref{fig:overview}, the structural reasoning process begins with the input image $\mathbf{I}_0$ and task instruction $l_{task}$. A Vision-Language Model (VLM) first identifies the task-relevant object to be manipulated. The corresponding object-centric visual observation, obtained via segmentation~\citep{SAM2}, is then passed to the Concept System.

Within the Concept System, a VLM-driven Concept Constructor~\citep{sun2025executable} organizes geometric primitives to instantiate a corresponding Manipulation Blueprint. This process composes reusable structural templates into an object-specific representation conditioned on the recognized category.

In parallel, the geometry-aware features extracted by VGGT are encoded into 3D-aware representations and fed into a Pre-trained Head~\citep{wei2026physicallygroundcommonsenseknowledge}, which serves as the Structural Parameter Head.  This module estimates the structural parameters $P_s$ and the initial kinematic state $P_k^0$ by fitting the instantiated blueprint to the observed object.

\subsection{Dynamic Parameter Tracking}

While the initial parameters $P_s$ and $P_k^0$ are precise, the kinematic parameters $P_k$ quickly become outdated as the robot interacts with the environment. 
Instead of relying on a separate, decoupled tracker, our approach directly leverages the rich, latent representations being learned by the main VLA model. As shown in Fig.~\ref{fig:overview}, we extract intermediate features ($\mathbf{F}_{\text{VLA}}$) from a specific \texttt{VLA Layer} (indicated by the dashed box). These features are channeled through a dedicated \texttt{Adapter} and a \texttt{Dynamic Parameter Head} to continuously estimate the change in the dynamic parameters, $P_k^t$.

\paragraph{VLA Feature Alignment}
To compensate for the VLA's inherent lack of robust 3D spatial understanding, we first employ external supervision signals to supervise its internal visual features~\citep{li2025spatialforcingimplicitspatial}.
We first input a set of multi-view RGB images $\boldsymbol{o}^\text{vis}_t$ into the pretrained 3D foundation model VGGT, which outputs the pixel-level spatial representations $\mathbf{F}_{\text{VGGT}}(\boldsymbol{o}^\text{vis}_t)$ . We then process the VLA's extracted intermediate features ($\mathbf{F}_{\text{VLA}}$) with batch normalization followed by an MLP-based Adapter to ensure compatibility in feature dimension with $\mathbf{F}_{\text{VGGT}}$.
The Alignment Strategy enforces consistency between the VGGT's spatial features and the VLA model's internal representation via a cosine similarity score. We define $\mathcal{L}_{align}$ to minimize the discrepancy between the VGGT features from the tracked images ($\mathbf{F}_{\text{VGGT}}$) and the adapted version of the VLA's internal feature vector ($\mathbf{F}_{\text{VLA}}$):
\begin{equation}
    \mathcal{L}_{align} = \mathbb{E}_{\boldsymbol{o}^\text{vis}} \left[ \mathcal{S}_{cos}(\text{Adapter}(\mathbf{F}_{\text{VLA}}),\, \mathbf{F}_{\text{VGGT}}) \right].
\end{equation}
By minimizing this loss, the VLA model is implicitly forced to learn features $\mathbf{F}_{\text{VLA}}$ that are predictive of the object's explicit structural state, thus internalizing the geometric and physical dynamics essential for the task.

\paragraph{Dynamic Parameter Head} 
The output of the Adapter, which aligns the VLA's latent space with structural features, is passed to the \texttt{Dynamic Parameter Head}. This head computes the updated dynamic parameter $P_k^t$ (e.g., $P_{\text{offset}}^t$), effectively tracking the object's kinematic state (such as a drawer's position) over time. Furthermore, to enhance the fine-grained spatial awareness of the parameter head, we utilize a cross-attention mechanism to embed object-centric 3D structure into the spatial representation. This infusion makes the head more sensitive to object boundaries and kinematic changes while avoiding the quadratic complexity associated with global attention on high-resolution maps.

This integration establishes a critical symbiotic loop: the VLA's powerful visual understanding aids in precise state tracking, while the Concept Expert's precise state tracking provides high-quality guidance back to the VLA. The dynamic tracker leverages the initial high-confidence parameters ($\mathbf{P}_0$) and the current visual observation ($\boldsymbol{o}^\text{vis}_t$) to hypothesize the updated state $\mathbf{P}_t$. Specifically, the VLA's own intermediate features ($\mathbf{F}_{\text{VLA}}$) or action predictions ($\boldsymbol{a}_t$) are used as a contextual prior to guide the tracker's attention, directing it particularly to regions of interest on the object.

\subsection{VLA Fine-Tuning}

The dynamically tracked Analytic Concepts ($\mathbf{P}_t$) serve as a powerful source of external supervision, significantly enhancing the precision and sample efficiency of the VLA fine-tuning process. The explicit nature of $\mathbf{P}_t$ allows us to define two primary mechanisms for Explicit Knowledge Injection: one providing direct supervision over action space and the other providing dense signals for reward optimization.


\paragraph{Kinematic Constraint Supervision}
To ensure high-precision physical interaction, we introduce Kinematic Constraint Supervision. It utilizes the structural integrity encoded in $\mathbf{P}_t$ to calculate the optimal 3D motion direction or force application required for manipulation.

The Concept Expert calculates the Optimal Interaction Reference Direction ($\boldsymbol{v}^*$), which is a ${3\text{D}}$ unit vector representing the instantaneous direction in which the robot should push or pull the object to follow its kinematic constraint (e.g., the precise linear sliding direction for a drawer, projected into the $3\text{D}$ action space). This ($\boldsymbol{v}^*$) acts as a direct supervisory signal on the VLA's action output.

We formulate $\mathcal{L}_{kcs}$ as a supervised loss term that minimizes the discrepancy between the VLA policy's predicted action direction $\boldsymbol{v}_{\boldsymbol{a}_t} = \frac{\boldsymbol{a}^{\text{trans}}_t}{||\boldsymbol{a}^{\text{trans}}_t||}$ and the ideal constraint vector $\boldsymbol{v}^*$ derived from $\mathbf{P}_t$:
\begin{equation}
\mathcal{L}_{\text{kcs}} = \mathbb{E}_{\boldsymbol{s}, \boldsymbol{v}^*} \left[ 1 - \frac{\boldsymbol{v}_{\boldsymbol{a}_t} \cdot \boldsymbol{v}^*_t}{\parallel \boldsymbol{v}_{\boldsymbol{a}_t} \parallel \cdot \parallel \boldsymbol{v}^*_t \parallel} \right]
\end{equation}

This mechanism directly maps the physical constraints encoded in the Analytic Concepts onto the VLA's action space, realizing direct supervision over precise interaction dynamics. This loss term is particularly effective when coupled with an \texttt{Action Expert} conditioned on $\boldsymbol{v}^*$.

\paragraph{Concept Derived Rewards}
For Reinforcement Learning (RL) fine-tuning, the explicit parameters in $\mathbf{P}_t$ enable the calculation of Concept Derived Rewards ($\mathcal{R}_\text{AC}$), overcoming the pervasive issue of reward sparsity in manipulation tasks. We decompose $\mathcal{R}_\text{AC}$ into two components: the Kinematic Progress Reward $\phi_{\text{prog}}$ and the {Affordance Alignment Reward} $\phi_{\text{afford}}$.

Kinematic Progress Reward leverages the continuously tracked kinematic parameters $P_{k}^t$ to define a differentiable measure of task completion progress. We define $\Delta P^t_k = P_k^t - P_k^0$ as the cumulative kinematic change, and $\Delta P^{\text{goal}}_k$ as the required change for task success. $\phi_\text{prog}$ is then computed based on the difference between the current kinematic change $\Delta P_k^t$ and the total required change $\Delta P_{k}^{\text{goal}}$:

\begin{equation}
   \phi_{\text{prog}}(\boldsymbol{s}_t) = \exp \left( - \frac{1}{\sigma} \parallel \Delta P_{k}^t - \Delta P_{k}^{\text{goal}} \parallel^2 \right), 
\end{equation}
where $\sigma$ controls the shape of the reward decay. It provides continuous, proximity-based feedback: the closer the current change is to the goal change, the higher the reward.

Affordance Alignment Reward utilizes the structural parameters $P_s$ and the Affordance Annotations embedded within the Analytic Concept assets. Let $\mathbf{G}_{\text{set}}$ be the set of all target Grasp Poses derived from the concept parameters $\mathbf{P}$. 
$\phi_{\text{afford}}$ quantifies the geometric alignment between the robot's end-effector pose $G_{\text{tcp}}$ and the nearest target affordance pose $G_{\text{target}}$ provided by the concept. This reward guides the initial phase of the task, ensuring the agent achieves a correct, physically viable pre-grasp or grasping configuration.
\begin{equation}
\phi_{\text{afford}}(\boldsymbol{s}_t) = \max_{G \in \mathbf{G}_{\text{set}}} \left( \exp \left( - \frac{1}{\rho} \cdot \text{Dist}(G_{\text{tcp}}, G) \right)\right),
\end{equation}
where $\rho$ is a scaling factor. The distance metric $\text{Dist}(G_1, G_2)$ is a combined $6\text{D}$ pose metric Each $6\text{D}$ pose, $\mathbf{G}$, is composed of a position $p \in \mathbb{R}^3$ and an orientation $q$. $\text{Dist}(\cdot)$ is defined as a weighted sum of the positional L2 norm and the angular error:
\begin{equation}
\text{Dist}(G_1, G_2) = \sqrt{w_p \cdot \parallel p_{1} - p_{2} \parallel^2 + w_q \cdot \theta_{\text{err}}^2}
\end{equation}
Here, $\theta_{\text{err}}$ is the angle (in radians) between the orientations $q_{\text{tcp}}$ and $q_{{G}}$, and $w_p, w_q$ are empirically determined weights used to balance the positional and angular alignment contributions. Our dense reward component $\mathcal{R}_\text{AC}$ is thus formulated by linearly combining these analytic feature functions:
\begin{equation}
    \mathcal{R}_\text{AC}(\boldsymbol{s}_t) = w_{\text{prog}} \cdot \phi_{\text{prog}}(\boldsymbol{s}_t) + w_{\text{afford}} \cdot \phi_{\text{afford}}(\boldsymbol{s}_t).
\end{equation}
This structure directly corresponds to the auxiliary part of the total reward function $r_t = \alpha \cdot I_{\text{success}} + (1 - \alpha) \cdot \mathcal{R}_\text{AC}(\boldsymbol{s}_t)$.



\paragraph{Total Optimization Objective} The complete objective for VLA fine-tuning under the SAGE framework combines the standard task loss ($\mathcal{L}_{task}$) with our explicit supervision loss ($\mathcal{L}_\text{kcs}$):
\begin{equation}
 \mathcal{L}_{total} = \mathcal{L}_{task} + \lambda_{k} \mathcal{L}_\text{kcs}  + \lambda_{a} \mathcal{L}_{align}
\end{equation}
Here, $\mathcal{L}_{task}$ is the task-specific loss (e.g., RL objective or Cross-Entropy for SFT), and $\lambda_{k}$ and $\lambda_{a}$ are the scalar hyperparameters. 
\begin{table*}[t]
  \centering
  \caption{\textbf{SimplerEnv simulation evaluation results for the Google Robot setup.} ``FT'' denotes performance of the  fine-tuned models. $^*$ marks the method specifically designed for 3D inputs.}
  \footnotesize
  \setlength{\tabcolsep}{1pt} 
\begin{tabular}{l*{8}{c}}
    \toprule
    \multirow{3}{*}{Model} & \multicolumn{3}{c}{\makecell{{Variant Aggregation}}} & 
    \multicolumn{3}{c}{\makecell{{Visual Matching}}} & \multirow{3}{*}{\textbf{Avg}} \\
    \cmidrule(lr){2-4} \cmidrule(lr){5-7} 
    &
    \makecell{Pick\\Coke Can} & 
    \makecell{Move \\ Near} & 
    \makecell{Open/Close\\  Drawer}  & 
    \makecell{{Pick}\\{ Coke Can}} & 
    \makecell{Move \\ Near} & 
    \makecell{{Open/Close}\\{ Drawer}} \\
    \midrule
    RT-1-X & 49.0\% & 32.3\% & 29.4\% & 56.7\% & 31.7\% & 59.7\% & 43.1\% \\
    Octo-Base  & 0.6\% & 3.1\% & 1.1\%  & 17.0\% & 4.2\% & 22.7\% &  8.11\% \\
    OpenVLA & 54.5\% & 47.7\% & 17.7\%  &16.3\% & 46.2\% & 35.6\% &  36.3\%\\
    RoboVLM & 68.3\% & 56.0\% & 8.5\%  & 72.7\% & 66.3\% & 26.8\% & 49.8\%  \\
    RoboVLM(FT) & 75.6\% & 60.0\% & 10.6\%  & 77.3\% & 61.7\% & 43.5\% & 54.8\%\\
    SpatialVLA(FT)$^*$ & \textbf{88.0\%} & 72.7\% & 41.8\%  & \textbf{86.0\%} & \textbf{77.9\%} & 57.4\% & 70.6\%  \\
    \midrule
    $\pi_0$ & {75.2\%} & 63.7\% & 25.6\%  & 72.7\% & 65.3\% & 38.3\% & 56.8\% \\
    $+$SAGE-SFT & \textbf{88.0\%} & 70.8\% & 33.3\%  & 76.4\% & 73.6\% & 55.6\% & 66.3\% \\
    \midrule
    OpenVLA-OFT & 65.3\% & 59.0\% & 12.2\%  & 72.3\% & 69.6\% & 47.2\% & 54.3\% \\
    $+$SAGE-SFT & 83.3\% & 70.8\% & 41.7\%  & 80.6\% & 75.0\% & 62.5\% & 69.0\%\\
    $+$SAGE-CQL & 84.7\% & \textbf{73.6\%} & \textbf{45.8\%} & 83.3\% & 76.3\% & \textbf{66.7\%} & \textbf{71.7\%}\\
    \bottomrule
    \end{tabular}%
  \label{tab:R2}%
\end{table*}%



\section{Experiment}

This section presents the empirical evaluation of the SAGE framework.  Additional details on model architecture, dataset construction, and training hyperparameters are provided in the Appendix~\ref{sec:setup}.

\subsection{Offline Learning Evaluation}

\paragraph{Experimental Context and Baselines}
To establish a fundamental comparison under controlled conditions, we conduct initial evaluations within the SimplerEnv~\citep{DBLP:conf/corl/LiHGMPWFLSKL0F024}, which features tasks with canonical kinematic constraints (e.g., linear sliding). As shown in Table~\ref{tab:R2}, our SAGE framework is built upon the OpenVLA-OFT model~\citep{OpenVLA-OFT}, which serves as our primary baseline. We benchmark this against a suite of standard VLA baselines, including $\pi_0$~\citep{, black2024pi0visionlanguageactionflowmodel}, Octo~\citep{ghosh2024octo}, OpenVLA~\citep{DBLP:conf/corl/KimPKXB0RFSVKBT24} and more concurrent works~\citep{qu2025spatialvla, DBLP:journals/corr/abs-2412-14058,DBLP:conf/rss/BrohanBCCDFGHHH23}, to provide a comprehensive performance landscape.

\paragraph{Performance Comparison}
We evaluate two distinct offline variants of our SAGE framework. The first, SAGE-SFT, assesses the power of Analytic Concept and Kinematic Constraint Supervision alone during standard SFT. The second, SAGE-CQL, represents the offline RL post-training model, integrating Concept-Derived Rewards ($\mathcal{R}_{\text{AC}}$) into the CQL objective~\citep{DBLP:conf/nips/KumarZTL20}. As demonstrated in Table 1, both SAGE variants significantly outperform the base model. SAGE-SFT significantly improves the average success rate of OpenVLA-OFT from 54.3\% to 69.0\%, and boosts $\pi_0$ from 56.8\% to 66.3\%. The SAGE-CQL model achieves the highest performance, reaching 71.7\% average, surpassing all other methods, including the  SpatialVLA(FT) baseline (70.6\%), which incorporates carefully designedspatial information embedding for 3D inputs. The impact of our approach is most evident in the kinematically complex ``Open/Close Drawer'' task. 

\begin{figure*}[tb]
  \centering
  \begin{subfigure}{0.3\textwidth}
  \centering
   \includegraphics[trim = 0mm 0mm 35mm 0mm, clip, width=0.99
  \linewidth]{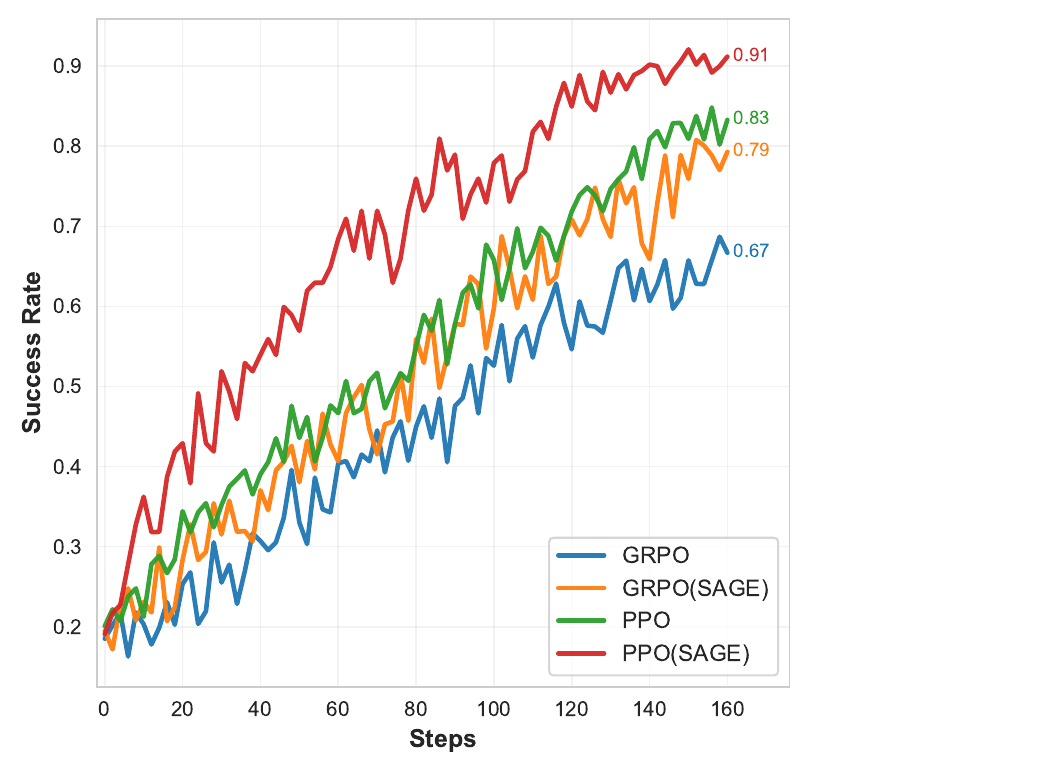}
  \end{subfigure}
  \hfill
\begin{subfigure}{0.3\textwidth}
  \centering
   \includegraphics[trim = 0mm 20mm 0mm 20mm, clip, width=0.99
  \linewidth]{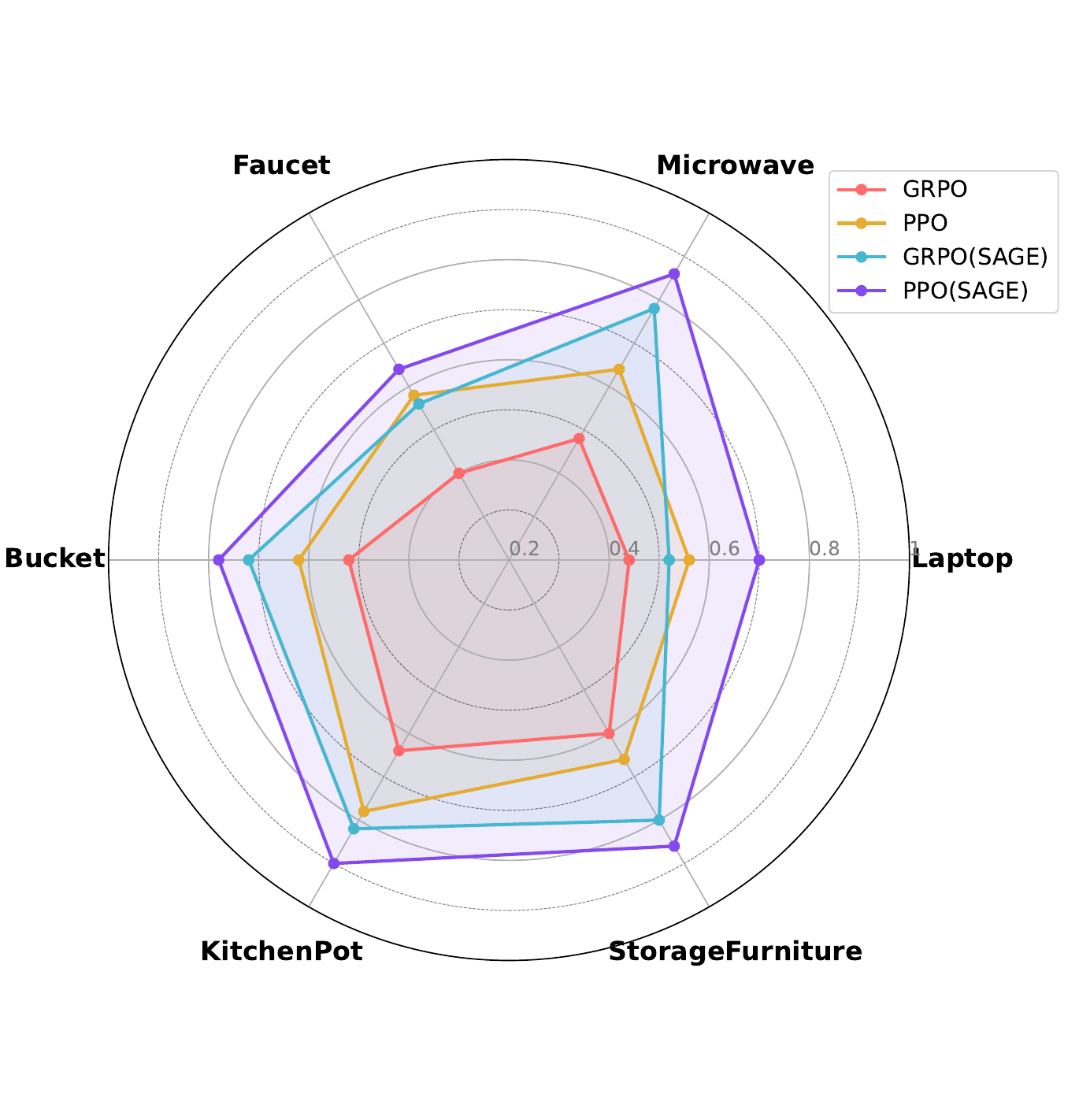}
  \end{subfigure}
  \hfill
  \begin{subfigure}{0.33\textwidth}
  \centering
   \includegraphics[trim = 0mm 0mm 20mm 0mm, clip, width=0.99
  \linewidth]{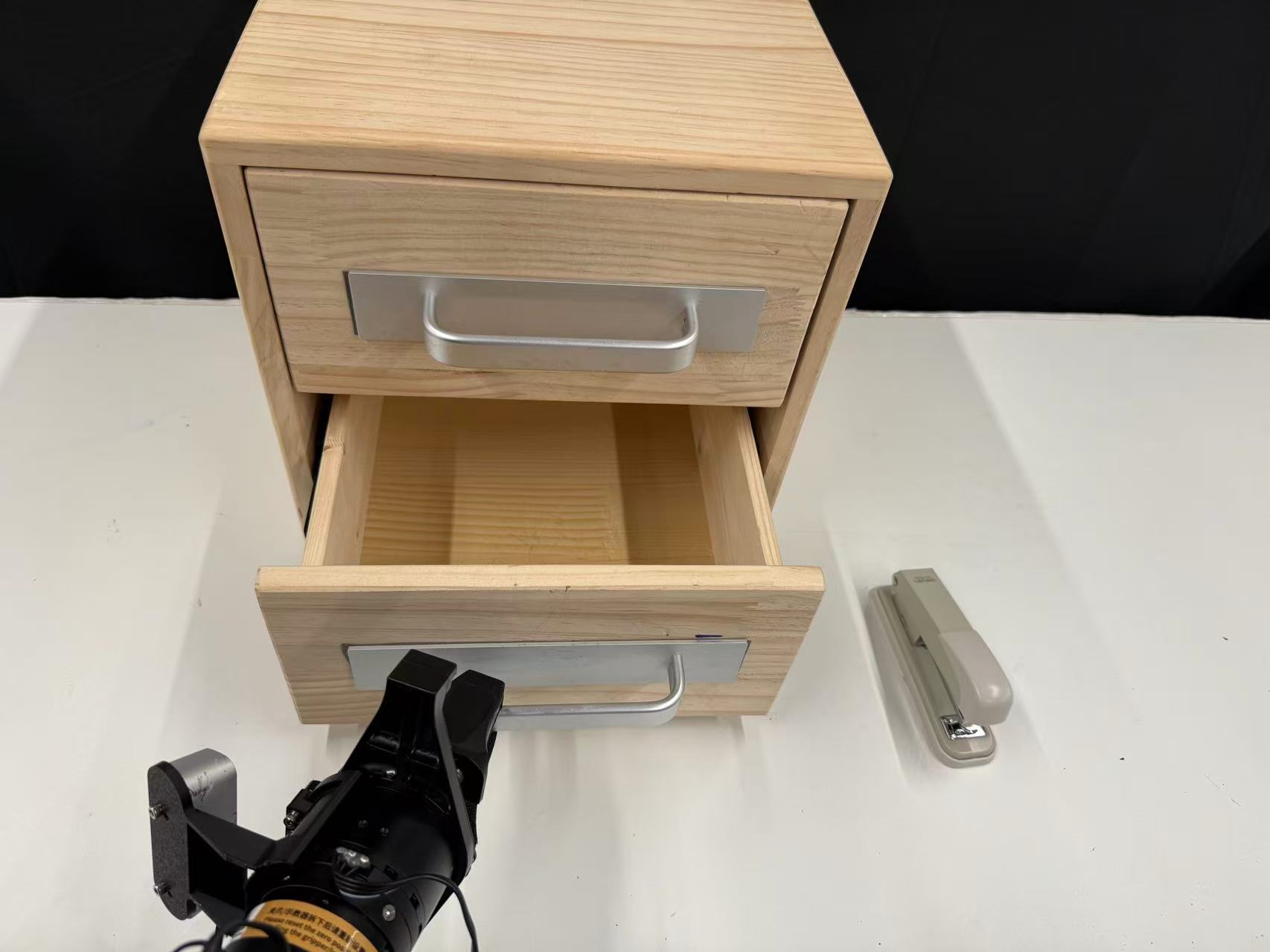}
   \end{subfigure}
    \caption{(a) Online RL performance comparison in the Open Drawer task. (b) The radar chart shows the online RL final success rates on manipulation tasks from the ShapeNet Mobility dataset. (c) A representative example of ``Place the stapler in the lower drawer'' task.}
  \label{fig:combined}
\end{figure*}

\subsection{Online Reinforcement Learning Performance}

\paragraph{Performance Comparison on Open Drawer task} To evaluate the capabilities of our framework during active interaction, we compare the SAGE method against established online reinforcement learning algorithms, including PPO and GRPO. This analysis focuses on the final success rate, the stability of the training process, and the overall data efficiency of SAGE when actively exploring the environment compared to purely reward-driven approaches. As shown in the learning curves (Fig.~\ref{fig:combined}(a)), the SAGE framework provides a clear and consistent advantage over its respective baselines on the Open Drawer task in SimplerEnv. The SAGE-enhanced variants demonstrate significantly faster convergence and achieve higher final success rates. The consistent gap between PPO(SAGE) and PPO, and GRPO(SAGE) and GRPO, indicates that the explicit knowledge injection provided by SAGE—combining the directional supervision and dense rewards—is highly effective at stabilizing exploration and accelerating the learning process in online settings.

\paragraph{Manipulation Task}

We extend the evaluation to a more complex environment tailored for articulated object manipulation. In the Sapien~\citep{DBLP:conf/cvpr/XiangQMXZLLJYWY20} simulation environment, we utilize asset data from the ShapeNet-Mobility dataset to construct six challenging manipulation tasks: Faucet, Microwave, Laptop, StorageFurniture, KitchenPot, and Bucket. For online RL training, we adopt a protocol where the VLA policy (OpenVLA-OFT) undergoes a warm-up phase of SFT, following RLinF~\citep{RLinf}. Detailed experimental settings are available in the supplementary material. We compare standard online RL baselines (PPO and GRPO) against their SAGE-enhanced counterparts. As illustrated in the radar plot (Fig.~\ref{fig:combined}(b)), the results demonstrate a clear and significant performance improvement from our SAGE framework. Both PPO(SAGE) and GRPO(SAGE) consistently outperform their non-SAGE baselines across all six task categories. Notably, PPO(SAGE) achieves the highest overall success rates, with near-perfect performance on tasks like ``Microwave'', ``StorageFurniture'', and ``KitchenPot''. This confirms that our explicit knowledge injection mechanisms are highly effective for online RL, enabling the agent to master diverse, contact-rich interactions where standard policies struggle.

\paragraph{Ablation Study.}

We conduct an ablation study on the Open Drawer task to evaluate the contribution of different components in SAGE. The results are reported in Table~\ref{tab:rw}(a). First, removing the alignment loss $\mathcal{L}_{align}$ consistently degrades performance, indicating that aligning the learned representations with analytic concepts contributes to stable policy learning, although it is not the primary source of the performance improvement.
Second, using the ground-truth parameters $P_k^{t*}$ leads to a slight performance gain. This suggests that accurate estimation of kinematic constraints plays a role in guiding interaction policies, while also indicating that the kinematic parameters inferred by SAGE are already sufficiently accurate for effective policy learning.
Finally, providing both ground-truth kinematic and structural parameters ($P_k^{t*}$ and $P_s^{*}$) results in comparable or only marginally improved performance, which implies that the structural parameters estimated by the VFMs are sufficiently precise, and the residual estimation error does not significantly affect the RL outcomes.



\subsection{Real-World Validation}

\begin{table}[tb]
\centering
\caption{Ablation study on the Open Drawer task and real-world manipulation performance.}
\label{tab:combined}

\setlength{\tabcolsep}{4pt} 
\renewcommand{\arraystretch}{0.95} 

\begin{subtable}{0.38\linewidth}
\centering
\footnotesize 
\caption{Ablation study}
\vspace{-2mm}
\begin{tabular}{lcc}
\toprule
Setting & PPO & GRPO \\
\midrule
SAGE & 0.91 & 0.79 \\
w/o $\mathcal{L}_{align}$ & 0.86 & 0.75 \\
w. $P_k^{t*}$ & 0.94 & 0.82 \\
w. $P_k^{t*}$ + $P_s^{*}$ & 0.93 & 0.84 \\
\bottomrule
\end{tabular}
\vspace{-2mm}
\label{tab:ab_transposed}
\end{subtable}
\hspace{2mm} 
\begin{subtable}{0.58\linewidth}
\centering
\footnotesize
\caption{Real-world evaluation}
\vspace{-2mm}
\begin{tabular}{ccc}
\toprule
Task & $\pi_{0.5}$ & +SAGE \\
\midrule
\makecell[l]{Place the stapler in the lower drawer.} & 60\% & 85\% \\
\makecell[l]{Place the cube in the kitchen pot.} & 75\% & 90\% \\
\makecell[l]{Place the bowl in the microwave.} & 50\% & 80\% \\
\makecell[l]{Place the green cube on the plate.} & 90\% & 100\% \\
\makecell[l]{Stack blue bowl on middle then stack the pink bowl.} & 60\% & 90\% \\
\bottomrule
\end{tabular}
\vspace{-2mm}
\label{tab:rw}
\end{subtable}

\end{table}
For the real-world evaluation, we utilize a AGILE PiPER Dual-Arm robot equipped with Orbbec DABAI cameras for visual input. We apply our SAGE-SFT framework to the baseline policy ($\pi_{0.5}$~\citep{intelligence2025pi05visionlanguageactionmodelopenworld}) and compare its performance against the unmodified $\pi_{0.5}$ policy on 5 challenging manipulation tasks. A representative example of these tasks is illustrated in Fig.~\ref{fig:combined}(c). The policies are evaluated based on their final success rate over 20 trials for each task. The results, presented in Table~\ref{tab:rw}, demonstrate that SAGE significantly enhances the policy's ability to execute precise, multi-step manipulation tasks in a real-world setting. See task descriptions in Appendix~\ref{sec:td}.

\section{Related Work}
\label{sec:relatedwork}

\paragraph{Vision-Language-Action Models} 
VLA models primarily achieve alignment between multimodal inputs and robot actions through imitation learning on large-scale datasets~\citep{DBLP:conf/corl/KimPKXB0RFSVKBT24, qu2025spatialvla, DBLP:journals/corr/abs-2412-14058, DBLP:conf/icml/ZhenQCY0DHG24, DBLP:journals/corr/abs-2510-13778, black2024pi0visionlanguageactionflowmodel,DBLP:conf/rss/BrohanBCCDFGHHH23,DBLP:conf/icra/TongDFWZCSZZDHL25,yu2025forcevla}. While pre-trained Vision-Language Models (VLMs) offer notable generalization capacity, supervised fine-tuning (SFT) is necessary to adapt them to specific tasks—for instance, by integrating lightweight adapters or enriching visual inputs with spatial information~\citep{goyal2025vla0buildingstateoftheartvlas,fan2025interleavevlaenhancingrobotmanipulation,li2025controlvlafewshotobjectcentricadaptation}. However, such SFT methods often lack adequate 3D geometric guidance, resulting in policies that struggle with high-precision alignment in constrained manipulation scenarios.  
Meanwhile, inspired by the success of Reinforcement Learning (RL) in fine-tuning Large Language Models (LLMs) and VLMs, recent efforts have explored RL to further improve VLA performance~\citep{li2025simplevlarlscalingvlatraining, liu2025rlbringvlageneralization,li2025vla}. Yet, these VLA-RL approaches are frequently limited by the difficulty of designing effective, non-sparse rewards for complex constrained tasks, hindering their real-world applicability.

\paragraph{Structural Representations for Manipulation.}
The operational paradigm of a manipulation system, governed by its internal representation, dictates its capabilities and limitations. While traditional approaches relying on complete rigid-body models excel in structured environments, their dependence on perfect a prior knowledge renders them fragile in the face of uncertainty~\citep{migimatsu2020object, dantam2018incremental}. The recent shift toward data-driven representations, e.g., spanning neural embeddings~\citep{hsu2023s,cheng2023nod,yuan2022sornet}, particle systems~\citep{bauer2024doughnet, abou2024physically}, and keypoints or descriptors~\citep{simeonov2022neural, manuelli2019kpam, huang2024rekep}, aims to overcome this brittleness. Nevertheless, these modern alternatives often introduce new challenges, such as computational instability, a need for extensive manual supervision, or latent geometric assumptions, which currently restrict their practical deployment. To bridge this gap, our work introduces analytic concepts~\citep{DBLP:conf/nips/0003LXWWZL24} that grounds the representation in physically meaningful structures.

\section{Conclusion}

We propose SAGE framework for VLA model post-training to perform complex manipulation by integrating explicit analytic concepts. Kinematic constraint supervision aligns the predicted action direction with the object’s physical trend. Concept derived rewards provide dense feedback based on kinematic progress and object grasp alignment. The comprehensive integration of the geometric and physical expertise fundamentally improves both SFT and RL convergence, achieving superior learning efficiency and robustness in manipulation tasks.


\medskip
\small
\bibliographystyle{ieeenat_fullname}
\bibliography{main}

\newpage
\appendix

\section{Implementation Details}
\label{sec:rationale}

\subsection{Analytic Concept System}\label{sec:ACS}

In this paper, we leverage the idea of Analytic Concept to facilitate more efficient annotation of 3D object knowledge by recognizing 3D objects through generalized concepts. And our purpose mainly focuses on how to leverage the rich knowledge into VLA post-training. The concept system in Fig. 2 has been validated for its effectiveness on a large collection of conceptualized data (\textbf{containing 40 categories and about 4,400 objects})~\citep{DBLP:conf/nips/0003LXWWZL24, sun2025executable}. However, it is worth emphasizing that Concept is not designed as a dataset for a specific task. The system can avoid certain human efforts involved in object conceptualization and then allow VLMs easily acquire a conceptual blueprints for objects in the manipulation task. 


\begin{table}[t]
\centering
\caption{
Quantitative evaluation of Analytic Concept instantiation.
The first row reports the average accuracy of concept identification by the VLM, measured in percentage.
The second row reports the average distance from the point cloud of the detected part to the mesh rendered from the estimated analytic concept parameters, measured in millimeters.
}
\label{tab:concept_reliability}
\resizebox{\linewidth}{!}{
\begin{tabular}{l|ccccccccccc|cccccc}
\toprule
\multirow{2}{*}{Evaluation}
& \multicolumn{11}{c|}{\textbf{Training Categories}}
& \multicolumn{6}{c}{\textbf{Testing Categories}} \\
\cmidrule(lr){2-12} \cmidrule(lr){13-18}
& Box & Dor & Fct & Fdr & Ket & Mcw & Stf & Swt & Tcn & Win & AVG
& Bkt & Pot & Saf & Tab & Wsm & AVG \\
\midrule
Accuracy
& 97.5 & 95.0 & 91.7 & 82.6 & 96.4 & 92.9 & 96.7 & 94.8 & 94.8 & 90.0 & 94.2
& 97.6 & 88.0 & 95.9 & 97.8 & 88.9 & 95.6 \\
Distance
& 9.24 & 3.66 & 1.92 & 1.85 & 2.65 & 5.67 & 5.42 & 1.84 & 9.80 & 8.62 & 5.18
& 9.73 & 3.41 & 9.75 & 8.65 & 12.4 & 6.55 \\
\bottomrule
\end{tabular}
}
\end{table}

\paragraph{Reliability of Analytic Concept Instantiation}
Since SAGE relies on the instantiated Analytic Concepts to provide structured guidance, we provide a quantitative evaluation of both concept identification and parameter estimation on PartNet-Mobility. The results are shown in Tab.~\ref{tab:concept_reliability}. 

The high concept identification accuracy (94.2\% on training categories and 95.6\% on testing categories) shows that the VLM-based Concept Constructor can reliably reason over Analytic Concept synopses, with only limited misidentifications. In addition, the parameterized nature of Analytic Concepts allows them to adapt to diverse object geometries. As a result, the system can tolerate certain concept identification errors: even when the VLM selects an imperfect concept due to recognition errors or hallucinations, the subsequent parameter estimation stage can still fit the concept to the observed geometry and recover structurally meaningful parameters. This suggests that our concept instantiation process is robust to occasional VLM errors and can provide reliable structural guidance for downstream policy learning.

\paragraph{Analytic Concepts' General Coverage on Objects}

To assess how many Analytic Concepts are needed to cover actionable parts in real-world articulated objects, we conduct an analysis over all 46 object categories in PartNet-Mobility. We randomly sample 20, 30, and 40 categories ten times and count the number of concepts required to cover their actionable parts. On average, these subsets require 27, 34, and 37 concepts, respectively, while covering all 46 categories requires only 39 concepts.
These results show a diminishing marginal increase in the number of required concepts as the number of object categories grows. This suggests that Analytic Concepts are not tied to isolated object categories, but instead capture reusable structural and manipulation patterns shared across categories. For example, the \texttt{L\_Handle} concept can be reused for doors, storage furniture, faucets, windows, and other categories. Such cross-category reuse supports the scalability of the concept library within articulated object domains and motivates its use as a structured interface for VLA fine-tuning.

\subsection{Network Architecture}
The static structural parameter estimation network takes the point cloud of the detected part with 2,048 points as input. The it utilizes a Point-Transformer~\cite{zhao2021point}  as encoder, which extracts 128 groups of points with size 32 from the input point cloud and has 12 6-headed attention layers. Subsequently, an average pooling layer is introduced to extract the global feature of the entire point cloud. Then, an MLP with three linear layers and accompanied ReLU activation outputs the structural parameters.
In Adapter for feature alignment, we use two layer MLP. Then the dynamic parameter head consists of 5 layer transformer with cross attention and 2 layer MLP.

\section{Experimental Setup}\label{sec:setup}

\paragraph{Offline Learning Setup}
The hyper-parameter SAGE:
\begin{itemize}
    \item We select GPT-4o as the VLM model for object parsing and concept selection.
    \item  The scalar hyperparameter $\lambda_{k} = 0.5$ balancing the contribution of the auxiliary constraint supervision.
    \item We use an annealed weight for the alignment loss:
$\lambda_a(t)=\lambda_a^0 \cdot \max\left(0, 1-\frac{t}{T_a}\right)$,
where \(\lambda_a^0=0.5\) and \(T_a\) denotes the alignment warm-up horizon. This schedule encourages the VLA representation to acquire spatially consistent features in the early stage of fine-tuning, while gradually reducing the constraint so that the policy can adapt to task-specific objectives.
    \item The total reward function weight $\alpha=0.8$. The concept reward weights $w_{\text{prog}} = 2$ and $w_{\text{prog}} = 1$. For convenience, $\sigma$ and $\rho$ are set to $1$.
\end{itemize}
We select CQL as the offline RL algorithm, which is an Offline RL algorithm that addresses extrapolation error by introducing a penalty to the Q-function loss. This penalty enforces the policy's expected Q-value to be an underestimate of the true return, primarily by minimizing Q-values for out-of-distribution (OOD) actions.

The Q-function loss includes a conservative term $\mathcal{L}_{\text{C}}(\theta)$:

$$
\mathcal{L}_{\text{CQL}}(\theta) = \mathbb{E}_{(s, a) \sim \mathcal{D}} \left[ \frac{1}{2} (Q_{\theta}(s, a) - y)^2 \right] + \alpha \cdot \mathcal{L}_{\text{C}}(\theta)
$$

The penalty term forces low Q-values generally, while ensuring high Q-values for actions present in the data:

$$
\mathcal{L}_{\text{C}}(\theta) = \mathbb{E}_{s \sim \mathcal{D}} \left[ \log \sum_{a} \exp(Q_{\theta}(s, a)) - \mathbb{E}_{a \sim \mathcal{D}(s)} \left[ Q_{\theta}(s, a) \right] \right]
$$

This conservatism ensures the policy relies only on actions reliably supported by the fixed training data. The policy is improved with SAC-style entropy regularization.

\paragraph{Online Reinforcement Learning Setup}
We follow RLinF~\citep{RLinf} to build our code and training pipeline (i.e., PPO and GRPO) and deploy on 8 NVIDIA H200 GPUs.

\paragraph{Manipulation Task} We conduct our experiment on 6 representative categories of objects. We would like our evaluation to reflect the VLA ability to understand articulated object structures and detect affordances on articulated object by reinforcement learning.  For each interaction
simulation, we initially place an object in the SAPIEN simulator at the center of the scene. The whole scene is observed through RGB-D cameras with known intrinsic parameters, which stares at the center of the object and is
positioned at Table similar to SimplerEnv. We consider an interaction
with a certain part successful if either of the following condition is hold: 1) absolute motion of the part is greater than 0.01. 2) motion of the part is greater than half of its maximum motion range.

\paragraph{Real World Experimental Setup}
For the critical real-world evaluation, we utilize the AGILE PiPER Dual-Arm robot platform, which serves as our physical execution environment. The robot is equipped with a high-resolution Orbbec DABAI camera mounted for comprehensive visual input.

Our fine-tuning baseline is the pre-trained Vision-Language-Action model, $\pi_{0.5}$~\citep{intelligence2025pi05visionlanguageactionmodelopenworld}, a powerful VLA backbone. We apply our SAGE-SFT framework specifically to adapt $\pi_{0.5}$ to our real-world constrained tasks.

To ensure a fair comparison and robust policy learning, we collected a small, task-specific dataset consisting of 25 expert demonstrations for each of the three evaluation tasks. The training process for the SAGE-SFT fine-tuning was performed using a distributed setup comprising 8 NVIDIA H200 GPUs, utilizing an aggregate batch size of $32 \times 8 = 256$. This high-throughput configuration allowed us to efficiently update the policy and leverage the single-shot nature of SFT while benefiting from the dense geometric supervision provided by SAGE.
\begin{figure}[tb]
  \centering
   \includegraphics[trim = 70mm 45mm 110mm 50mm, clip, width=0.4
  \linewidth]{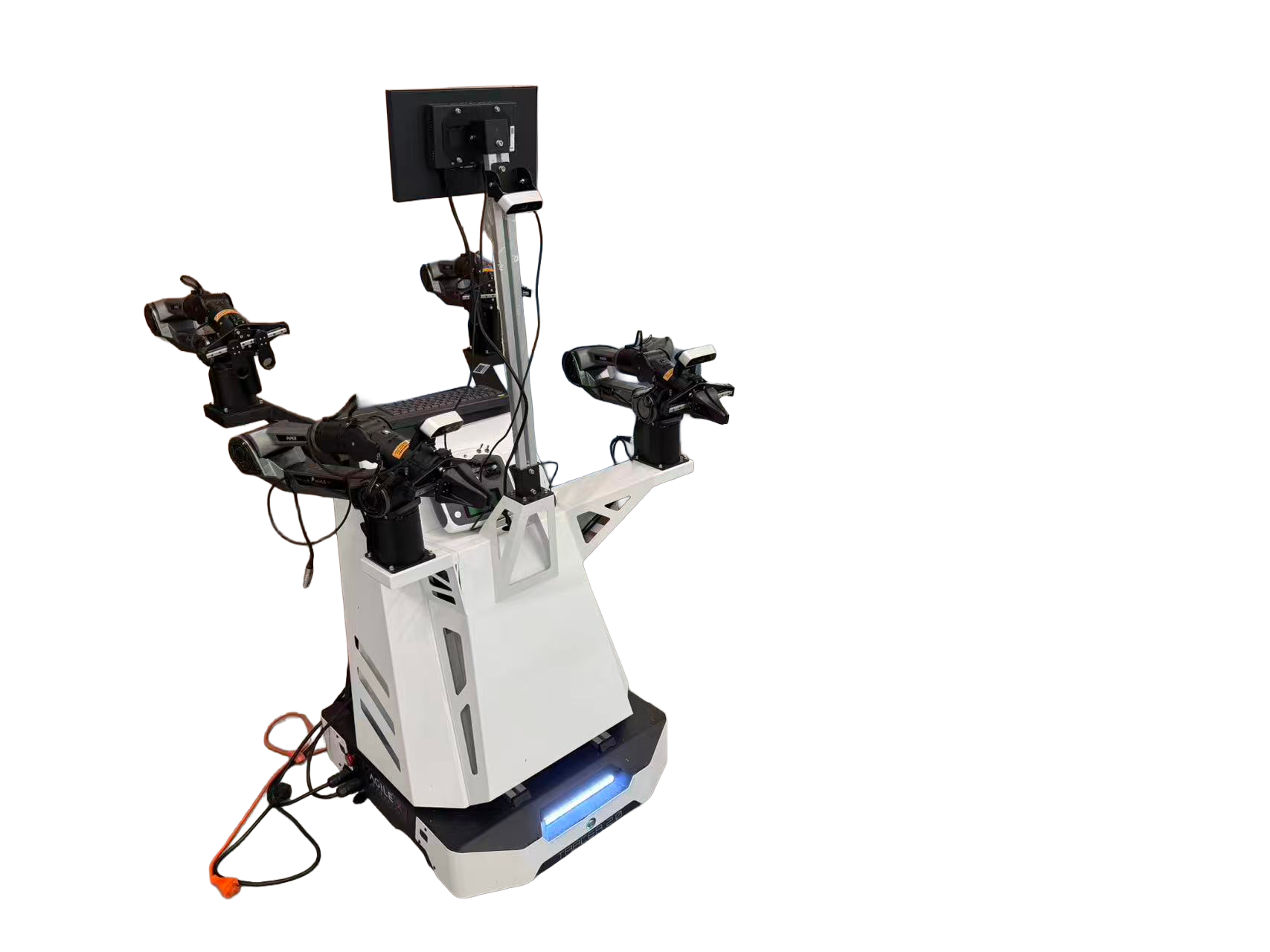}
   \caption{Hardware Configuration.}
   \label{fig:hw}
\end{figure}

\section{Task Description}\label{sec:td}

\textbf{1. Place the stapler in the lower drawer:} 
In this task, the robot is required to open or interact with a drawer and place a stapler into the lower compartment. The task involves articulated object manipulation and object placement within an enclosed space.

\textbf{2. Place the cube in the kitchen pot:} 
This task requires the robot to place a cube into a kitchen pot. The manipulation involves positioning the object relative to a container with a defined opening.

\textbf{3. Place the bowl in the microwave:} 
In this task, the robot is required to place a bowl inside a microwave. The task involves interacting with an articulated object and placing the target object into its interior space.

\textbf{4. Place green cube on plate:} In this experiment, we introduced cubes of different colors as distractors to evaluate the model's ability to focus on the target object (green cube) while ignoring visually similar but task-irrelevant objects. This tests the model's robustness to perceptual distractions and its precision in target identification.

\textbf{5. Stack blue bowl on middle then stack the pink bowl:} This sequential manipulation task was designed to assess the model's generalization capability regarding spatial positions. The robot must first stack the blue bowl onto the middle-positioned bowl, and then stack the pink bowl onto the resulting stack. Successful execution requires understanding of relative spatial relationships, sequential task planning, and adaptation to varying stacking positions, thereby validating the model's spatial generalization ability.

The execution videos are included in the supplementary materials.

\section{Limitations and Future Work}
SAGE has several limitations that point to future research directions. First, our framework relies on the Analytic Concept System to provide executable structural and manipulation blueprints. Although Analytic Concepts are reusable across object categories and can be extended compositionally, the current system is mainly designed for objects with clear structural regularities, such as articulated objects. Extending Analytic Concepts to deformable objects is an important direction for future development.

Second, the blueprint instantiation process depends on external perception modules, including VLM-based object identification, segmentation, and VGGT-based 3D feature extraction for static parameter estimation. These modules introduces extra computation during concept initialization, with an average inference time of approximately 2.1 seconds in our implementation. However, this cost is incurred only during the concept initialization stage and does not affect the inference latency of the VLA policy during execution.

Third, while SAGE is designed as a modular framework for VLA fine-tuning, our current experiments focus on a limited set of VLA architectures. In future work, we plan to evaluate SAGE on a broader range of VLA backbones to further assess its generality and practical deployment potential.

\clearpage

\newpage

\end{document}